\documentclass[a4paper,conference]{IEEEtran}

\usepackage{subfigure}
\usepackage{url}
\usepackage{times}
\usepackage{amssymb}
\usepackage{array}
\usepackage{multirow}
\usepackage[ruled,lined]{algorithm2e}
\usepackage{amsmath}
\usepackage{booktabs}
\usepackage{cite}

\ifCLASSINFOpdf
  \usepackage[pdftex]{graphicx}
  \graphicspath{{../pdf/}{../jpeg/}}
  \DeclareGraphicsExtensions{.pdf,.jpeg,.png}
\else
  \usepackage[dvips]{graphicx}
  \graphicspath{{../eps/}}
  \DeclareGraphicsExtensions{.eps}
\fi

\begin{document}
\title{Temporal Action Detection by Joint Identification-Verification\\}
\author{
\IEEEauthorblockN{Wen Wang }
\IEEEauthorblockA{
University of Electronic Science\\and Technology of China\\
2006 Xiyuan Ave.\\
Chengdu, Sichuan, 611731\\
Email: uestc\_wangwen@163.com
}
\and
\IEEEauthorblockN{Yongjian Wu}
\IEEEauthorblockA{
State Key Laboratory of Synthetical\\Automation for Process Industries\\Northeastern University\\
Shenyang, Wenhua Road, 110819\\
Email: wyj\_neu@163.com
}
\and
\IEEEauthorblockN{ Haijun Liu, Shiguang Wang, Jian Cheng }
\IEEEauthorblockA{
University of Electronic Science\\and Technology of China\\
2006 Xiyuan Ave.\\
Chengdu, Sichuan, 611731\\
Email: haijun\_liu@126.com
}

}

\maketitle
\begin{abstract}
Temporal action detection aims at not only recognizing action category but also detecting start time and end time for each action instance in an untrimmed video. The key challenge of this task is to accurately classify the action and determine the temporal boundaries of each action instance. In temporal action detection benchmark: THUMOS 2014, large variations exist in the same action category while many similarities exist in different action categories, which always limit the performance of temporal action detection. To address this problem, we propose to use joint Identification-Verification network to reduce the intra-action variations and enlarge inter-action differences. The joint Identification-Verification network is a siamese network based on 3D ConvNets, which can simultaneously predict the action categories and the similarity scores for the input pairs of video proposal segments. Extensive experimental results on the challenging THUMOS 2014 dataset demonstrate the effectiveness of our proposed method compared to the existing state-of-art methods for temporal action detection in untrimmed videos.
\end{abstract}
\IEEEpeerreviewmaketitle

\section{Introduction}

\begin{figure}
  \centering
  \includegraphics[width=9cm]{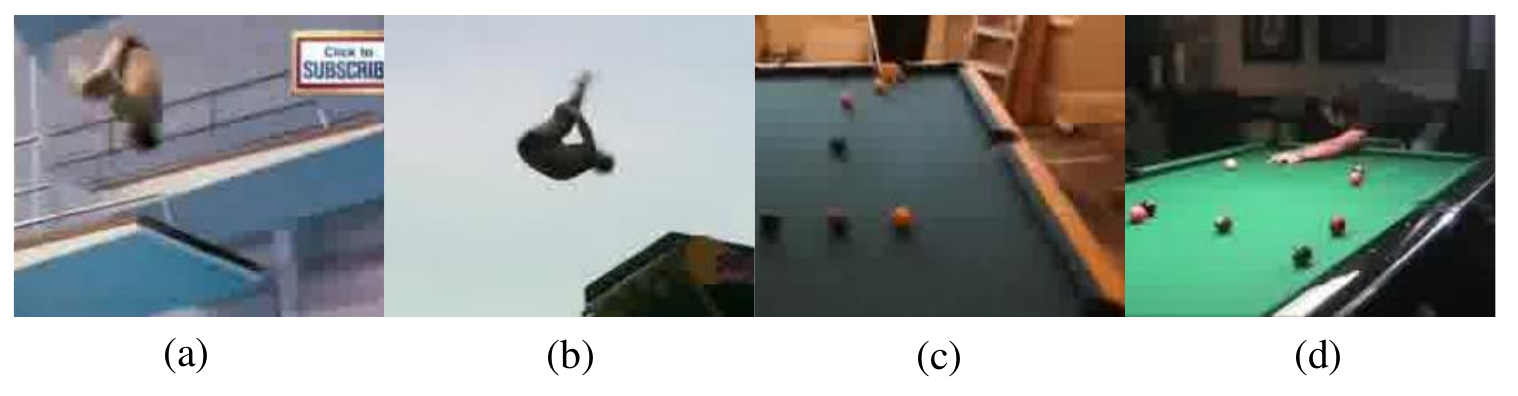}
  \caption{Samples from THUMOS 2014 dataset: (a) Diving. (b) CliffDiving. (c) CricketBowling. (d) CricketBowling. (a) and (b) are from different action labels, but they perform more similarities, (c) and (d) are from the same action label of CricketBowling, while they perform more variations due to different background.}
  \label{fig:Samples_img}
\end{figure}

Temporal action detection has risen much attention in recent years because of the continuously booming of videos in the Internet. This task is a very challenging problem, given a long untrimmed video, action detection aims to predict the action categories and also localize the start and end time of actions of interest. However, actions of the same label could look much different in different poses, duration, background and so on. At the same time, actions of different classes may show many similarities. Such variations within the same action class and similarities within different action classes could make action detection more challenging, especially in untrimmed videos. There are multiple activities per video and many similarities between different action classes in the large-scale video dataset THUMOS 2014 \cite{THUMOS14}, like \emph{CliffDiving} and \emph{Diving} as shown in Fig. \ref{fig:Samples_img} (a) and (b). There are  also large variations in pose of the same action class, like \emph{CricketBowling} as shown in Fig. \ref{fig:Samples_img} (c) and (d).  Therefore, it is an essential topic in temporal action detection task to reduce the intra-action variations while enlarge the inter-action differences.

Current methods \cite{scnn_shou_wang_chang_cvpr16,Zhu2016Efficient,Xu2017R,Shou2017CDC} for temporal action detection always adopt the proposal-classification framework, which has been very successful in object detection. This proposal-classification framework consists of three steps. Firstly, action proposals are generated as candidates by temporal sliding windows. Secondly, the candidate segments are used to train a classifier for action recognition. Finally, some post-processing and non-maximum suppression (NMS) procedures are conducted to refine temporal boundaries from proposal segments to precisely localize boundaries of action instances. Instead of sliding window, \cite{Zhao2017Temporal} used a method called TAG \cite{Xiong2017A} to generate proposals. SSAD \cite{Lin2017Single} directly detect action instances in untrimmed videos by proposing a novel Single Shot Action Detector network to skip the proposal generation step. However, these methods ignore the problem of low classification accuracy which is caused by intra-action variations and inter-action differences.

In this work, we also adopt the proposal-classification framework to perform temporal action detection task. However, our work only focuses on action classification. For temporal action classification, we argue that it is effective to use identification and verification simultaneously for reducing intra-action variations while enlarging inter-action differences. Given a set of video proposal segments, identification network can output the category of each video proposal segments, which can be treated as a multi-class recognition task \cite{Yi2014Deep,Wu2016PersonNet,Varior2016Gated}. While verification network takes a pair of video proposal segments as input and determine whether they belong to the same action or not, which can be treated as a binary-class classification task. The verification model ignores the relationship between the input video proposal pairs with other proposals, and the identification model also dosen't take similarity between video proposal pairs into consideration. To address this problem, our work proposes to combine identification and verification model into a siamese network that can not only predict action categories but also estimate the similarities of the input video proposal pairs.

Our contributions are two-fold:

(1) To the best of our knowledge, our work is the first to incorporate the identification and verification model into temporal action detection framework. We propose a Identification-Verification siamese (IVS) network that predicts action categories and similarity scores at the same time, thus improving temporal action detection accuracy.

(2) Our proposed method significantly outperforms the state-of-art methods on the challenging action detection dataset: THUMOS 2014.

The paper is organized as follows. We first review some related works in Section II. In Section III, we describe our framework of Identification-Verification. Section IV provides the experimental results on the large-scale action detection dataset. At last, we conclude our works in Section V.

\section{Related work}
In this section, we will briefly review some related works, including action recognition, object detection, temporal action detection and person Re-identification.

\noindent {\bf Action recognition.}
Action recognition has been made great progress in the past few years with the wide adoption of convolutional neural networks (CNNs). Action recognition aims to determine the category of an action instance, so the models of action recognition can be used in temporal action detection to extract features. The previous works \cite{Simonyan2014Two,Weinzaepfel2015Learning,Wang2015Towards} try to capture both appearance and motion information, both of which are essential to action recognition. However, most of these methods are based on the short trimmed videos, in which the only action is performed during the whole video time. Hense, there is no need to consider the temporal boundaries of each action instance. Typical datasets used in action recognition such as UCF101 \cite{Soomro2012UCF101}, HMDB51 \cite{Kuehne2013HMDB51} and Sports-1M \cite{Karpathy2014Large} include amount of trimmed videos.

\noindent {\bf Object detection.}
Earlier methods of object detection are mostly based on hand-crafted low-level features. In recent years, inspired by the success of deep learning, object detection has achieved state-of-art performance. Most approaches are based on the pipeline of proposal-classification. R-CNN \cite{Ross2013Rich} uses SelectiveSearch \cite{Uijlings2013Selective} for proposal generation, CNN framework for feature extraction, SVM for proposal classification, and bounding box for regression. Fast R-CNN \cite{Girshick2015Fast} uses RoI pooling layer to make features update all network layers. Faster R-CNN \cite{Ren2015Faster} uses RPN network to generate region proposals which reduces the time consumption. In our work, we also adopt the proposal-classification pipeline for temporal action detection task.

\noindent {\bf Temporal action detection.}
Temporal action detection framework gets closely related to object detection framework. This task focuses on learning how to detect action instances in untrimmed videos where the boundaries and categories of action instances have been annotated. Most works focus on the framework of detection which are based on proposal and classification. For example, SCNN \cite{scnn_shou_wang_chang_cvpr16} uses sliding windows to generate proposals, and then classifies proposal segments of different categories. To get temporal boundaries of each action instance, SCNN uses localization network to boost temporal localization accuracy. SSN \cite{Zhao2017Temporal} tries to determine the completeness of each action instance via a structured temporal pyramid, and this can effectively improve the accuracy of classification and localization. However SSAD \cite{Lin2017Single} dosen't use the proposal generation step, which directly detects action instances and predicts confidence scores of action instances in untrimmed videos.

\noindent {\bf Person re-identification.}
Person re-identification (re-ID) has become increasingly popular due to its application and research significance \cite{Zheng2016Person}. It is usually regarded as an image retrieval problem and aimed at determining whether the person has been appeared in another camera. Many works \cite{Chen2014Deep,Chopra2005Learning,Zheng2016A} used verification models to person re-identification. Verification models usually take a pair of images as input and output a similarity score by calculating the distance between features. Our works are different from these. Firstly, we jointly use identification and verification in temporal action detection task to improve the accuracy of classification. Secondly, our model is pretrained on 3D ConvNets (C3D) \cite{Xu20133D,Tran2014Learning} which has shown great performance in video analysis \cite{Xu2017R}. Thirdly, our network is trained to minimize the joint of identification and verification losses.

\section{Our approach}

\begin{figure*}
\centering
\includegraphics[width=15cm]{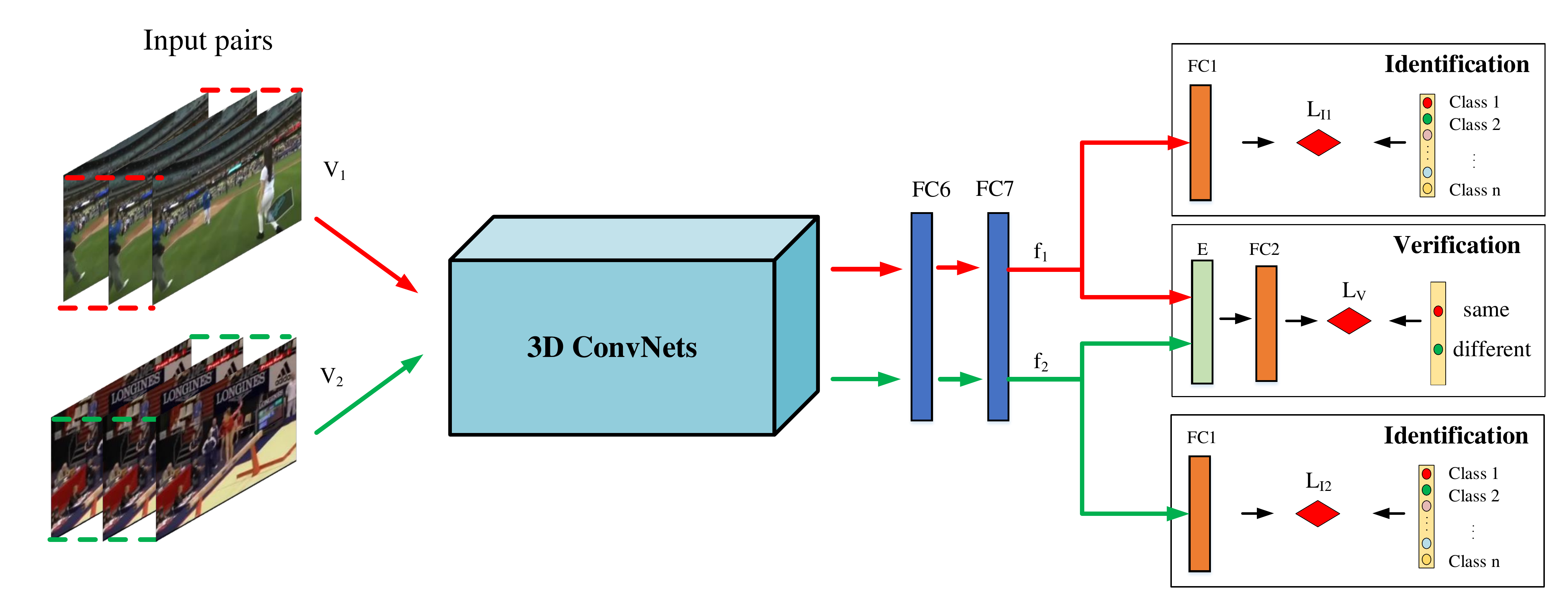}
\caption{The structure of our siamese network: A pair of action proposals of size ($16 \times 112 \times 112$, denoting length $\times$ height $\times$ width), are fed to 3D ConvNets for feature extraction. The red and the green video data streams share weights on 3D ConvNets and output 4096-dim feature vectors ${f_{1}}$, ${f_{2}}$. The two features are respectively fed into our identification model for action classification and simultaneously fed into our verification model for similarity estimation. The siamese network is only used during training stage. }
\label{fig:framework_sim}
\end{figure*}

In this section, we will introduce our Identification-Verification siamese (IVS) approach in details. The framework of our approach is shown in Fig. \ref{fig:framework_sim}. We apply a siamese network which jointly trains identification and verification models. We use 3D ConvNets as our basic network architecture and simultaneously consider identification loss for action category prediction and verification loss for similarity estimation.

\subsection{Siamese network}
Our siamese network combines identification model and verification model. Given an input pair of action proposals $V_1$ and $V_2$, and their action labels, the two inputs share weights on 3D ConvNets, and we get two high-level features ${f_{1}}$, ${f_{2}}$ after two fully connected layers. Then each of them is fed to our identification model for action prediction. Simultaneously, ${f_{1}}$ and ${f_{2}}$ are also fed to our verification model for similarity estimation.

\begin{figure}[h]
  \centering
  \includegraphics[width=9cm]{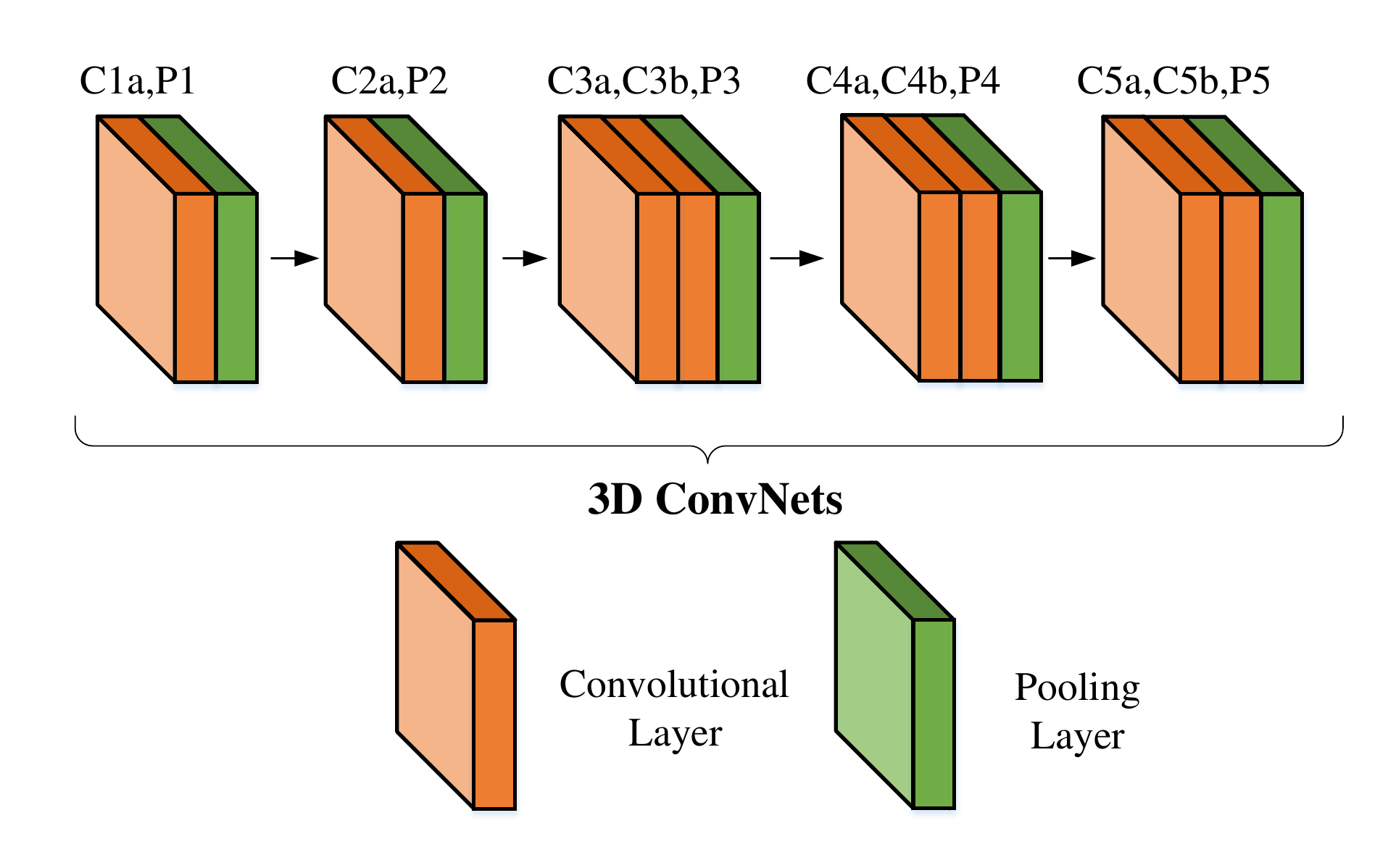}
  \caption{The architecture of 3D ConvNets.}
  \label{fig:C3D_network}
\end{figure}

\noindent {\bf 3D ConvNets.} Our siamese network adopts 3D ConvNets as the basic architecture for extracting high-level video action features. The 3D ConvNets consist of five layers. Each layer contains one or two 3D Convolutional Layer and one Pooling Layer. We use the notations \emph{C} (number of filter) for the 3D Convolutional Layer and \emph{P} (temporal kernel size, temporal stride) for the 3D Pooling Layer. The layout of these two types of layer in our architecture are shown in Fig. \ref{fig:C3D_network} is as follows: \emph{ C1a(64) - P1(1,1) - C2a(128) - P2(2,2) - C3a(256) - C3b(256) - P3(2,2) - C4a(512) - C4b(512) - P4(2,2) - C5a(512) - C5b(512) - P5(2,2)}.

\noindent {\bf Identification model.} After 3D ConvNets, there are two fully connected layers and both of them are with 4096 filters. For identification model, we add \emph{FC1} and the softmax to produce a distribution over the ${n}$ action categories. There are $21$ action classes in THUMOS 2014 dataset.

\noindent {\bf Verification model.} For verification model, we add an additional \emph{E} layer after the two fully connected layers of \emph{FC6} and \emph{FC7} for similarity estimation. The \emph{E} layer computes the euclidean distance of the two high-level feature ${f_{1}}$ and ${f_{2}}$. After that, we add a fully connected layer \emph{FC2} and the output is fed to a $2$-way softmax which produces a distribution over the $2$ class (same / different).

\subsection{Loss Function}
In our siamese network, we not only determine the action labels of the input pair-wise video proposal segments, but also estimate the similarity which we called the verification signal. The network is trained to minimize the three losses, two identification losses and one verification loss.

\noindent \textbf{Identification loss.} For identification model, we predict the action label for each input video segment, through a $softmax$ operation to predict the probability distribution $\hat{p} = [\hat{p}_1,\hat{p}_2,\cdots,\hat{p}_n]$ over the $n$ action classes,
\begin{align}
\hat{p} = softmax(\alpha _{I} * f),
\end{align}
where $f$ is the high-level action feature, ${\alpha _{I}}$ denotes the parameters of \emph{FC1}, and $*$ denotes the convolutional operation.

Our identification model calculates the loss between the predicted probability $\hat{p}$ and the truth probability $p$,
\begin{align}
L_{I}(\hat{p},p)=\sum_{i=1}^{n}-p_{i} \log(\hat{p}_{i}), \label{eq:fztsc}
\end{align}
where the truth probability $p = [p_1,p_2,\cdots,p_n]$ is defined,
\begin{align}
p_i=
\left\{
    \begin{tabular}{cl}
    $1$, & if it is the $i$th action,\\
    $0$, & otherwise.
    \end{tabular}
    \right.
\end{align}

\begin{figure*}
\centering
\includegraphics[width=18cm]{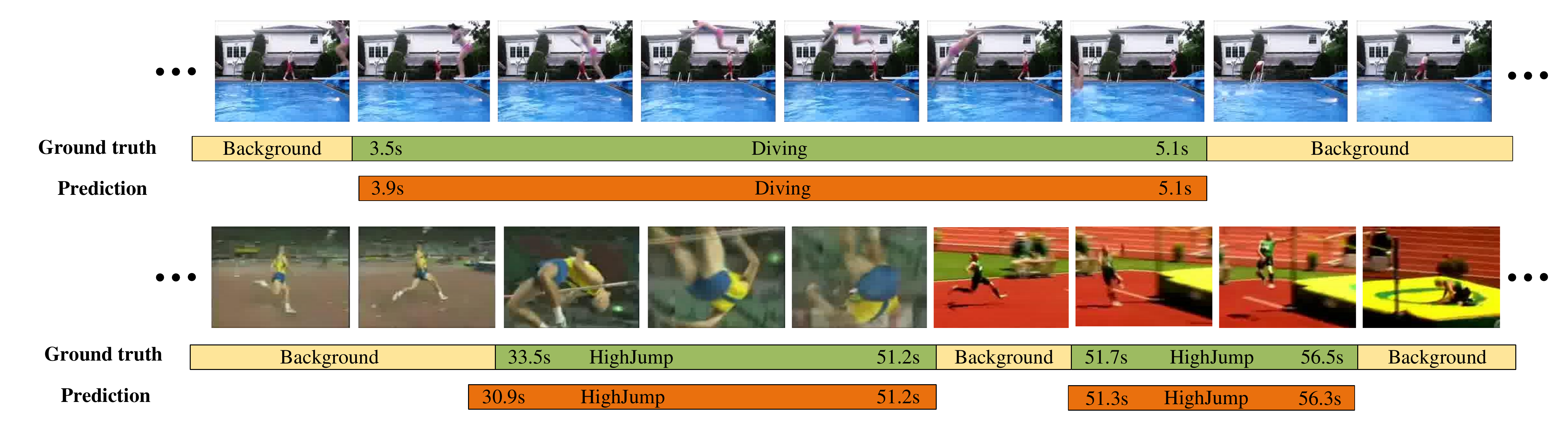}
\caption{The prediction results of our proposed method for two action instances in THUMOS 2014 test set. Our prediction results are always in accordance with the ground truth.}
\label{fig:instances_predict}
\end{figure*}

\noindent \textbf{Verification loss.}
To get high classification accuracy, we introduce the thought of verification to encourage the features extracted from videos of the same action class to be similar and enlarge the variations of features extracted from the different actions at the same time.
We add a verification signal (similarity score) $s = [s_1, s_2] = [0, 1]$, where $1$ means the two video proposal segments are from the same action category, while $0$ denotes different action categories.

Firstly, we compute the Euclidean distance $f_E$ of the two high-level features $f_1$ and $f_2$ through the $E$ layer, $f_E = \sqrt{(f_1 - f_2)^2}$.

Then the $softmax$ operation is adopted to predict the similarity score $\widetilde{p} = [\widetilde{p}_1, \widetilde{p}_2]$ over $2$ classes (different / same),
\begin{align}
\widetilde{p} = softmax(\alpha _{V} * f_E),
\end{align}
where $\alpha _{V}$ denotes the parameters of \emph{FC2}.

Finally, following the computation of identification loss, our verification loss is defined between the verification signal $s$ and the predicted similarity score $\widetilde{p}$,
\begin{align}
L_{V}(s,\widetilde{p}) = \sum_{i=1}^{2}-s_{i} \log(\widetilde{p}_{i}).
\end{align}

\noindent \textbf{Overall loss.}
In our siamese network, given an input pair of action proposals $V_1$ and $V_2$, the overall loss is formed by combining identification losses and verification loss,
\begin{align}
L = L_{I1} + L_{I2} + \lambda L_{V},
\end{align}
where $L_{I1}$ and $L_{I2}$ are the identification loss corresponding to the action proposals $V_1$ and $V_2$ respectively. $L_{V}$ is the verification loss. $\lambda \geq 0$ is a tradeoff parameter to balance the contribution of the two parts, and through empirical validation, we find that $\lambda = 1 $ works well in practice.

\section{Experiments}
In this section, we evaluated our proposed method on standard benchmark dataset: THUMOS 2014. We first introduce the evaluation dataset and the experimental settings, then compare our proposed Identification-Verification siamese (IVS) network with some state-of-art approaches.

\subsection{Dataset}
The temporal action detection THUMOS 2014 dataset is challenging and widely used. The dataset contains two tasks: action recognition and temporal action detection. For action detection, the dataset contains $1010$ videos for validation and $1574$ videos for testing, both of which are untrimmed videos. For action recognition, the training set is UCF101 of short trimmed videos with no temporal annotations. Following the protocol in \cite{scnn_shou_wang_chang_cvpr16,Xu2017R}, we used $200$ validation set videos (including $3007$ action instances) and $213$ test set videos (including $3358$ action instances) with temporal annotation to train and evaluate our Identification-Verification network.

\subsection{Evaluation metrics}
The THUMOS 2014 dataset has its own convention of reporting performance metrics, which evaluates Average Precision (AP) for all action categories and calculate mean Average Precision (mAP) for evaluation. We just follow the convention and calculate the mAP at different IoU thresholds. A prediction is correct only when it has the correct category prediction and its temporal IoU with ground truth instance is larger than the overlap threshold. The IoU thresholds used to compare the performances in our experiment are $0.1, 0.2, 0.3, 0.4, 0.5$.

\subsection{Implementation details}
We implemented our network based on Caffe \cite{Jia2014Caffe}. For temporal action detection, we used $200$ annotated untrimmed video in THUMOS 2014 validation set as our training set. Firstly, we directly used ActivityNet dataset \cite{Heilbron2015ActivityNet} to finetune the R-C3D model \cite{Xu2017R} to generate the training and testing proposals. Secondly, we used C3D model \cite{Xu20133D} parameters to initialize our Identification-Verification network, and trained it with the generated training proposals. We used SGD to learn CNN parameters in our framework, with batch size 5 and momentum $0.9$. The initial learning rates were set to $0.001$ for our siamese network.

\subsection{Experimental results}
We conducted extensive experiments to demonstrate the effectiveness of our proposed Identification-Verification siamese network for temporal action detection, including the following aspects.

\begin{figure*}
\centering
\includegraphics[width=18cm]{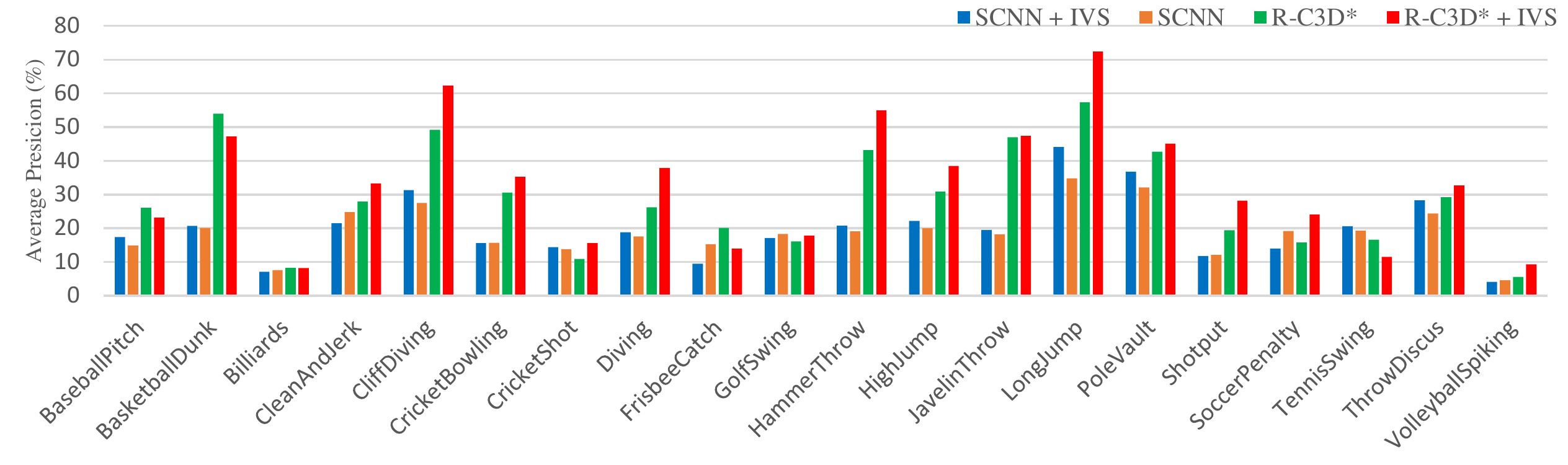}
\caption{The detection AP results over different action categories with overlap IoU threshold ${0.5}$ on THUMOS 2014 action detection dataset.}
\label{fig:ap_result}
\end{figure*}

\noindent {\bf Comparison with state-of-art methods.}
We compared our IVS method with six state-of-art methods, including PSDF \cite{Yuan2016Temporal}, SCNN \cite{scnn_shou_wang_chang_cvpr16}, CDC \cite{Shou2017CDC}, SSAD \cite{Lin2017Single}, R-C3D \cite{Xu2017R} and SSN \cite{Zhao2017Temporal}. The results  are shown in TABLE \ref{table_1}. The results of those comparison methods are from the original papers. From TABLE \ref{table_1}, we can see that our IVS method achieves much better performance than the compared methods especially when the overlap IoU are $0.3$, $0.4$ and $0.5$. Our IVS method only performs worse than SSN when the overlap IoU are $0.1$ and $0.2$. However, the performance under low overlap IoU is always less important compared to the performance under high overlap IoU in detection task.
We also show two prediction results of our proposed IVS method for the THUMOS 2014 test set in Fig. \ref{fig:instances_predict}. It shows the excellent performance of our IVS method compared to the ground truth.

\begin{table}[!htbp]
\centering
\caption{The mAP results of our method and other state-of-art methods on THUMOS 2014 with various IoU threshold $\theta $ used in evaluation. - indicates that results are not available in the corresponding papers.}
\begin{tabular}{p{2.5cm}<{\centering}p{0.75cm}<{\centering}p{0.75cm}<{\centering}p{0.75cm}<{\centering}p{0.75cm}<{\centering}p{0.75cm}<{\centering}}
\toprule
 $\theta $ & 0.5 & 0.4 & 0.3 & 0.2 & 0.1 \\
\midrule
PSDF \cite{Yuan2016Temporal} &  18.8 & 26.1 & 33.6 & 42.6 & 51.4\\
SCNN \cite{scnn_shou_wang_chang_cvpr16} & 19.0 & 28.7 & 36.3 &  43.5 & 47.7\\
CDC \cite{Shou2017CDC} &  23.3 & 29.4 & 40.1 &  - & -\\
SSAD \cite{Lin2017Single} &  24.6 & 35.0 & 43.0 &  47.8 & 50.1\\
R-C3D \cite{Xu2017R} &  28.9 & 35.6 & 44.8 &  51.5 & 54.5\\
SSN \cite{Zhao2017Temporal} &  29.8 & 41.0 & 51.9 &  {\bf 59.4} & {\bf 66.0}\\
\midrule IVS (ours) & {\bf 32.2} & {\bf 45.4} & {\bf 54.1} &  58.3 &   60.0\\
\bottomrule
\end{tabular}
\label{table_1}
\end{table}

\noindent {\bf The effect of proposals.}
For Identification-Verification siamese network, we adopted the SCNN \cite{scnn_shou_wang_chang_cvpr16} and R-C3D \cite{Xu2017R} as our proposals generation.
For SCNN, we directly put the results of SCNN as proposals to our IVS network, which is termed as SCNN + IVS. For R-C3D, we used ActivityNet \cite{Heilbron2015ActivityNet} dataset to finetune the model which is different from the original paper that was finetuned with UCF101 \cite{Soomro2012UCF101} dataset. We term the method of finetuned with ActivityNet dataset as R-C3D$^*$. Then the results of R-C3D$^*$, as proposals, were directly fed to ours IVS network, which is termed as R-C3D$^*$ + IVS.
As shown in TABLE \ref{table_2}, we can draw the following conclusions. (1) Compared R-C3D$^*$ to R-C3D, the results show that finetuned with ActivityNet can obtain better performance than finetuned with UCF101. (2) Our proposed IVS network can be used to improve the performance of previous works, demonstrated by the comparison of R-C3D$^*$ + IVS to R-C3D$^*$ and SCNN + IVS to SCNN. (3) Better proposal generations always achieve better detection results. R-C3D$^*$ performs better than SCNN denoting that R-C3D$^*$ is a better proposal generations compared to SCNN. This leads to the better performance of R-C3D$^*$ + IVS compared to SCNN + IVS.
The corresponding detection AP results for each category of SCNN, R-C3D$^*$ and our method with overlap IoU threshold ${0.5}$ are shown in Fig. \ref{fig:ap_result}.

\begin{table}
\centering
\caption{The mAP results of SCNN, R-C3D and our methods on THUMOS 2014 with various IoU threshold $\theta $ used in evaluation. We report the results using the same evaluation metrics as in \cite{scnn_shou_wang_chang_cvpr16}. }
\begin{tabular}{p{2.5cm}<{\centering}p{0.75cm}<{\centering}p{0.75cm}<{\centering}p{0.75cm}<{\centering}p{0.75cm}<{\centering}p{0.75cm}<{\centering}}
\toprule
 $\theta $ & 0.5 & 0.4 & 0.3 & 0.2 & 0.1 \\
\midrule
SCNN \cite{scnn_shou_wang_chang_cvpr16} & 19.0  & 28.7 & 36.3 &  43.5 & {\bf47.7} \\ \midrule
SCNN + IVS & {\bf 19.9 } & {\bf 31.2 } & {\bf 39.8} & {\bf 45.1} &   {\bf47.7}\\
\midrule \midrule
R-C3D \cite{Xu2017R} &  28.9 & 35.6 & 44.8 &  51.5 & 54.5 \\ \midrule
R-C3D$^*$  &  30.2 & 41.3 & 48.5 &  53.4 & 56.2\\
\midrule
R-C3D$^*$ + IVS & {\bf 32.2} & {\bf 45.4} & {\bf 54.1} & {\bf 58.3} &  {\bf 60.0}\\
\bottomrule
\end{tabular}
\label{table_2}
\end{table}

\noindent {\bf Parameter $\lambda$.}
$\lambda$ is a tradeoff parameter to balance the contribution of identification and verification. To explore the effects of identification and verification,  we conducted experiments with different values $\lambda = \{0, 0.5, 1, 2\}$. As shown in TABLE \ref{table_4}, comparing  ${\lambda = 0}$ with ${\lambda = 0.5}$, we can see that verification loss bring improvement for temporal action detection.  We can get the best mAP results when $\lambda = 1$. The results show that our joint Identification-Verification achieves good performance for action detection.

\begin{table}
\centering
\caption{The mAP results of our proposed method with different parameter $\lambda$.  }
\begin{tabular}{p{2.5cm}<{\centering}p{0.75cm}<{\centering}p{0.75cm}<{\centering}p{0.75cm}<{\centering}p{0.75cm}<{\centering}p{0.75cm}<{\centering}}
\toprule
 $\theta $ & {0.5} & {0.4} & {0.3} & {0.2} & {0.1} \\
\midrule
$\lambda = 0 $  & 18.5  & 28.0 & 36.2 &  41.4 & 47.1\\
\midrule
$\lambda = 0.5 $ & 19.8 & 30.5 & 39.6 &  44.8 &  47.4\\
\midrule$\lambda = 1 $  & {\bf 19.9} & {\bf 31.2 } & {\bf 39.8} &  {\bf 45.1} & {\bf 47.7}\\
\midrule
$\lambda = 2 $  & 18.7  &  28.3 & 37.8 & 42.7 &  45.0\\
\bottomrule
\end{tabular}
\label{table_4}
\end{table}

\noindent {\bf Verification loss vs. Contrastive loss\cite{Hadsell2006Dimensionality}.}
We adopted the verification loss to measure the similarity of the input video proposal pairs. To study the effect of our verification loss, we conducted experiment by replacing it with the contrastive loss which is a commonly used pair-similarity measure in siamese network. Since the entire model was trained end-to-end with jointly optimize identification loss and verification loss, we only replaced the verification loss by contrastive loss in our overall loss function. As shown in TABLE \ref{table_5}, we find that using contrastive loss leads to lower mAP results than our verification loss, which demonstrates the effectiveness of our verification loss.

\begin{table}
\centering
\caption{Comparison of verification loss and contrastive loss on our siamese network. The mAP results of the different $\theta $ are listed}.
\begin{tabular}{p{2.5cm}<{\centering}p{0.75cm}<{\centering}p{0.75cm}<{\centering}p{0.75cm}<{\centering}p{0.75cm}<{\centering}p{0.75cm}<{\centering}}
\toprule
 $\theta $ & {0.5} & {0.4} & {0.3} & {0.2} & {0.1} \\
\midrule
$Verification $ $loss$  & {\bf 19.8} & {\bf 30.5} &  {\bf 39.6} & {\bf 44.8} &  {\bf 47.4}\\
\midrule$Contrastive$ $loss$   & { 18.8}& { 29.8} & { 37.9} &  { 41.6} & { 43.6}\\
\bottomrule
\end{tabular}
\label{table_5}
\end{table}

\section{Conclusion}
In this paper, we propose a Identification-Verification siamese network to get high classification accuracy by reducing intra-action differences and enlarging inter-action variances. The proposed model simultaneously considers identification for action classification and verification for similarity estimation. Experimental results on THUMOS 2014 dataset demonstrate the effectiveness of our proposed siamese network for temporal action detection. Our joint Identification-Verification model also can be easily applied on different proposal generation networks, and achieve significantly improvements compared to other methods in the temporal action detection task.


\bibliographystyle{IEEEtran}

\bibliography{reference}

\end{document}